\documentclass[runningheads]{llncs}

\usepackage{eccv}

\usepackage{eccvabbrv}

\usepackage{graphicx}
\usepackage{booktabs}

\usepackage[accsupp]{axessibility}

\usepackage{hyperref}

\newlength\savewidth\newcommand\shline{\noalign{\global\savewidth\arrayrulewidth
  \global\arrayrulewidth 1pt}\hline\noalign{\global\arrayrulewidth\savewidth}}

\usepackage{array}
\newcolumntype{C}[1]{>{\centering\arraybackslash}m{#1}}
\newcolumntype{R}[1]{>{\raggedleft\arraybackslash}m{#1}}
\newcolumntype{P}[1]{>{\raggedright\arraybackslash}p{#1}}
\newcolumntype{M}[1]{>{\centering\arraybackslash}m{#1}}
\usepackage{multirow}

\usepackage{orcidlink}

\begin{document}

\title{Unleashing the Potential of the Semantic Latent Space in Diffusion Models for Image Dehazing} 

\titlerunning{Abbreviated paper title}

\author{Zizheng Yang \and Hu Yu \and Bing Li \and Jinghao Zhang \and Jie Huang \and Feng Zhao\thanks{Corresponding Author.}}

\authorrunning{Yang et al.}

\institute{MoE Key Laboratory of Brain-inspired Intelligent Perception and Cognition, \\
University of Science and Technology of China \\
\email{\{yzz6000, yuhu520, bing0123, jhaozhang, hj0117\}@mail.ustc.edu.cn},\\ 
\email{fzhao956@ustc.edu.cn} }

\maketitle

\begin{abstract}
Diffusion models have recently been investigated as powerful generative solvers for image dehazing, owing to their remarkable capability to model the data distribution.
However, the massive computational burden imposed by the retraining of diffusion models, coupled with the extensive sampling steps during the inference, limit the broader application of diffusion models in image dehazing.
To address these issues, we explore the properties of hazy images in the semantic latent space of frozen pre-trained diffusion models, and propose a Diffusion Latent Inspired network for Image Dehazing, dubbed DiffLI$^2$D.
Specifically, we first reveal that the semantic latent space of pre-trained diffusion models can represent the content and haze characteristics of hazy images, as the diffusion time-step changes. Building upon this insight, we integrate the diffusion latent representations at different time-steps into a delicately designed dehazing network to provide instructions for image dehazing. 
Our DiffLI$^2$D avoids re-training diffusion models and iterative sampling process by effectively utilizing the informative representations derived from the pre-trained diffusion models, which also offers a novel perspective for introducing diffusion models to image dehazing.
Extensive experiments on multiple datasets demonstrate that the proposed method achieves superior performance to existing image dehazing methods.
Code is available at \url{https://github.com/aaaasan111/difflid}.

\keywords{Image Dehazing \and Diffusion Models\and Latent Space}
\end{abstract}

\section{Introduction}
\label{sec:intro}

Image dehazing aims to recover a clean image from its hazy counterpart, which is critical to high-level vision tasks such as image classification~\cite{he2016deep, krizhevsky2017imagenet, simonyan2014very} and object detection~\cite{he2017mask, lin2017focal, ren2015faster}. It is challenging and ill-posed due to the infinite possible solutions for a given hazy image. Conventional methods utilize the physical scattering model~\cite{mccartney1976optics} to estimate the clean images. With the development of deep learning, convolution neural network (CNN) and Transformer-based methods have achieved great success in image dehazing~\cite{yu2022frequency, chen2019pms, zhang2018densely, zamir2022restormer, wang2022uformer, liang2021swinir}.
Recently, diffusion models~\cite{ho2020denoising, song2020score} exhibit great impressive performance in image generation, and achieve unprecedented success in downstream tasks, such as image editing~\cite{hertz2022prompt, brooks2023instructpix2pix} and personalization~\cite{ruiz2023dreambooth, gal2022image, liu2023cones}. Meanwhile, the diffusion models also significantly broaden the scope of possibilities for image dehazing.

The prevailing approach to applying diffusion models for image dehazing is to re-train a diffusion model that is conditioned on the hazy image from scratch~\cite{luo2023image, yue2023resshift}. These methods utilize the hazy image as the condition, and concatenate it with the noise map, which aims to implicitly guide the diffusion models to predict the corresponding clean image during the reverse process. Such paradigm requires re-training the entire diffusion models, which typically costs massive time and computation resources. On the other hand, the potential time-consuming sampling in the reverse process also limits their application.

\begin{figure*}[t!]
    \centering
    \includegraphics[width=0.98\linewidth]{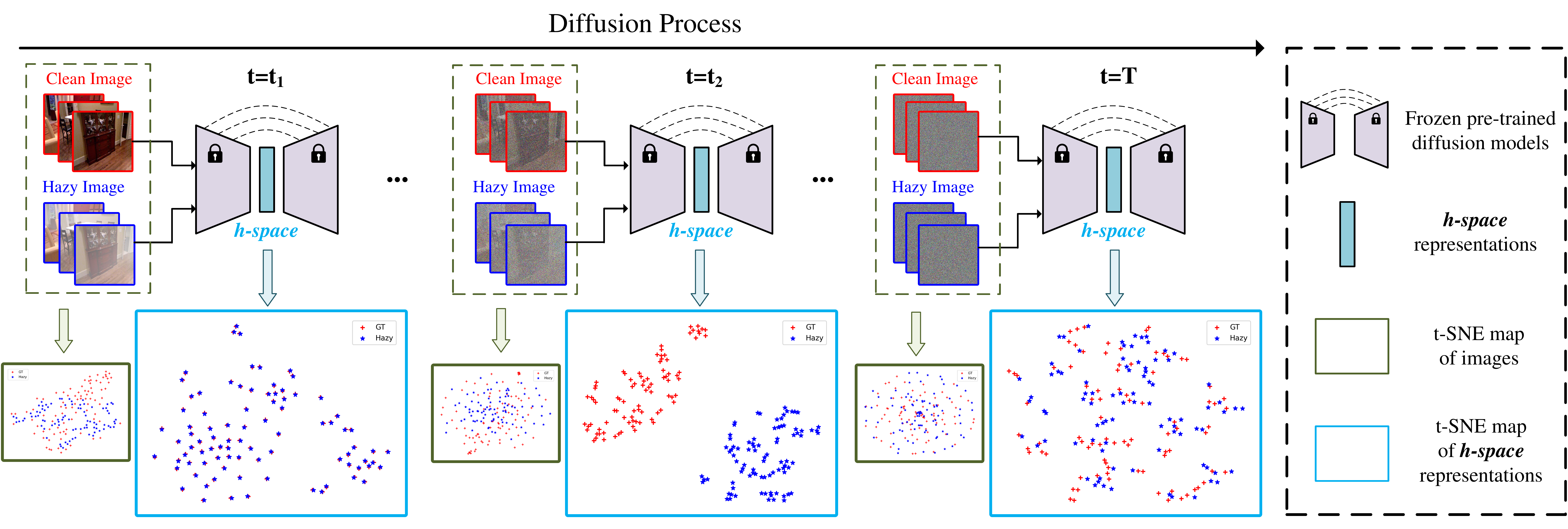}
    \vspace{-2mm}
    \caption{Distributions of hazy images and their corresponding clean images in \emph{h-space} and image space at different time-step $t$ during the diffusion process. When the time-step $t$ is small (\ie, $t=t_1$), the \emph{h-space} features with the same underlying image content are tightly clustered together, while those with different image contents are separated. When $t$ is large (\ie, $t=t_2$), the \emph{h-space} features of hazy and clean images are distinguished. Note that when $t$ becomes too large (\ie, $t=T$), the distribution of hazy and clean \emph{h-space} features becomes chaotic and irregular. The t-SNE maps in image space are also presented for comparison. Please zoom in for better view.}
    \label{Fig: Motivation}
\end{figure*}

To address the above issues, we try to investigate the potential of diffusion models for image dehazing from a new perspective: \emph{``Can we directly leverage the rich knowledge contained in pre-trained diffusion models, instead of re-training diffusion models from scratch?''} To this end, we investigate the properties of hazy images within the semantic latent space of frozen pre-trained diffusion models. Previous works~\cite{kwon2022diffusion, park2024understanding, jeong2024training} have discovered that the semantic latent space (named \emph{h-space}) has nice properties for high-level semantic manipulation. It is essential to investigate whether the \emph{h-space} also exhibits properties necessary for low-level image dehazing.
Specifically, we discover that, as the diffusion time-step changes, the \emph{h-space} representations of hazy images undergo a gradual transformation, transitioning from primarily encoding the underlying image contents to increasingly capturing the haze characteristics. Fig.~\ref{Fig: Motivation} describes the properties and Sec.~\ref{Sec: H-Space} provides detailed analysis.
Note that our exploration of \emph{h-space} for image dehazing is different from previous works that focus on the intermediate outputs during the diffusion process. To the best of our knowledge, it is the first attempt to explore the potential of the semantic latent space in pre-trained diffusion models towards image dehazing.

The aforementioned observation promotes us to leverage the informative \emph{h-space} representations to facilitate image dehazing. To this end, we propose a new framework, called the Diffusion Latent Inspired network for Image Dehazing (DiffLI$^2$D), which aims to integrate the \emph{h-space} representations for effective image dehazing. The DiffLI$^2$D adopts a hierarchical architecture similar to U-Net~\cite{ronneberger2015u}, enabling it to learn multi-scale features for image dehazing~\cite{zamir2022restormer, wang2022uformer, xia2023diffir}.
Specifically, to facilitate the content recovery of hazy images, we design a content integration module (CIM), which assists the DiffLI$^2$D in restoring image contents by utilizing the content representations derived from \emph{h-space}.
Furthermore, for better haze removal, a haze-aware enhancement (HAE) module is developed. It leverages the haze representations obtained from \emph{h-space} as guidance, enabling DiffLI$^2$D to remove the haze from the input hazy images effectively.

Moreover, the proposed DiffLI$^2$D does not require re-training any diffusion models, and circumvents the time-consuming reverse sampling process. Compared with existing diffusion model-based methods~\cite{xia2023diffir, yi2023diff}, our DiffLI$^2$D costs less computation resources.

We summarize our main contributions as follows:

\begin{itemize}

\item To the best of our knowledge, this is the first attempt to explore the \emph{h-space} of diffusion models for image dehazing. Additionally, we propose the DiffLI$^2$D framework for image dehazing through leveraging the informative representations derived from \emph{h-space}.

\item Our findings reveal a transition in the \emph{h-space} representations of hazy images, shifting from encoding the image contents to capturing haze characteristics, as the diffusion time-step changes.

\item Considering the properties of \emph{h-space} representations, we develop two modules, namely CIM and HAE, to facilitate the content recovery and haze removal in DiffLI$^2$D by leveraging the features derived from \emph{h-space}.

\item Extensive experiments demonstrate the superiority of our method. Moreover, the DiffLI$^2$D requires less computation resources, since it avoids re-training diffusion models and time-consuming reverse sampling process.

\end{itemize}

\section{Related Work}

\subsection{Image Dehazing}
Image dehazing aims to recover a clean image from its hazy version. Conventional approaches use physical scattering model~\cite{mccartney1976optics}, and try to regularize the solution space with various image priors~\cite{berman2016non, fattal2008single, he2010single}. However, these hand-crafted image priors may not be reliable. Recently, deep learning-based methods have dominated the image dehazing algorithms~\cite{yu2022frequency, chen2019pms, zhang2018densely, yang2022self, zheng2023curricular, qin2020ffa, ye2022perceiving}. For example, AOD-Net~\cite{li2017aod} tries to recover the clean images by reformulating the physical scattering model. AECR-Net~\cite{wu2021contrastive} introduces contrastive regularization to image dehazing. FSDGN~\cite{yu2022frequency} attempts to recover clean images through both spatial and frequency domains.
Recently, transformer~\cite{vaswani2017attention} is also introduced to image dehazing task and has achieved great success~\cite{qiu2023mb, zamir2022restormer, wang2022uformer, liang2021swinir, chen2021pre}.

\subsection{Diffusion Models}
Diffusion model~\cite{ho2020denoising, sohl2015deep}, as a newly emerged generative model, has achieved remarkable progress in image generation~\cite{dhariwal2021diffusion} and various downstream tasks, like image editing~\cite{hertz2022prompt, brooks2023instructpix2pix} and personalization~\cite{ruiz2023dreambooth, gal2022image, liu2023cones, kumari2023multi}.
Taking the DDPM~\cite{ho2020denoising} as an example, it constructs a Markov chain, and trains a denoising network, which aims to accurately fit target distributions.
Current diffusion models-based image restoration methods can be divided into two categories.
The first one is to re-train a diffusion model from scratch~\cite{yue2023resshift, luo2023image, saharia2022image, yi2023diff}, which often demands massive computation resources and time.
The second one is to guide the pre-trained diffusion models to generate target images by constraining the reverse sampling~\cite{chung2022diffusion, song2022pseudoinverse}, avoiding re-training diffusion models. However, the time-consuming reverse sampling and the need for accurate degradation process limit their applications. The work~\cite{kwon2022diffusion} explores the properties of the semantic latent space in pre-trained diffusion models (\ie, \emph{h-space}) for high-level semantic manipulation. Despite this, the characteristics of \emph{h-space} for low-level image restoration are yet to be explored.

\section{Preliminary: Diffusion Models}

In this paper, we follow the DDPM~\cite{ho2020denoising}, and briefly introduce the key points in diffusion models. Concretely, it consists of a $T$-steps forward process that gradually adds Gaussian noise to the input image $x_0$, and a reverse process that learns to generate images by progressively denoising. 

In the forward process, for any $t \in [0, T]$, we can get the current state $x_t$:

\begin{equation}
\label{Eq: Diffusion forward}
    q(x_t|x_{t-1}) = \mathcal{N}(x_t;\sqrt{1 - \beta_t}x_{t-1}, \beta_t\mathbf{I}),
\end{equation}
where $x_t$ is the noisy image at time-step $t$, $\beta_t$ is the variance schedule~\cite{ho2020denoising}, and $\mathbf{I}$ is the identity matrix. Through the reparameterization, we can get $x_t$ given $x_0$:

\begin{equation}
\label{Eq: Diffusion forward 2}
    q(x_t|x_0)=\mathcal{N}(x_t;\sqrt{\bar{\alpha}_t}x_0, (1 - \bar{\alpha}_t)\mathbf{I})),
\end{equation}
where $\alpha_t = 1 - \beta_t$, and $\bar{\alpha}_t = \prod_{s=0}^{t}\alpha_s$.

During the reverse process, the diffusion models aim to estimate the previous state $x_{t-1}$ from the current state $x_t$. We can get the posterior distribution $p(x_{t-1}|x_t, x_0)$ through the Bayes' theorem:

\begin{equation}
\label{Eq: Diffusion reverse}
    p(x_{t-1}|x_t, x_0) = \mathcal{N}(x_{t-1};\mu_t(x_t, x_0), \sigma_t^{2}\mathbf{I}),
\end{equation}
where the mean $\mu_t(x_t, x_0) = \frac{1}{\sqrt{\alpha_t}}(x_t - \frac{1 - \alpha_t}{\sqrt{1 - \bar{\alpha}_t}}\epsilon)$, and the variance $\sigma_t^2 = \frac{1 - \bar{\alpha}_{t-1}}{1 - \bar{\alpha}_t}\beta_t$. DDPM leverages a neural network $\epsilon_{\theta}$ to estimate the noise $\epsilon$ in $\mu_t(x_t, x_0)$. For any time-step $t \in [0, T]$, we can get the loss function defined in~\cite{ho2020denoising}:

\begin{equation}
\label{Eq: Diffusion loss function}
    L(\theta) = \| \epsilon - \epsilon_{\theta}(\sqrt{\bar{\alpha}_t}x_0 + \sqrt{1 - \bar{\alpha}_t}\epsilon) \|_2^2.
\end{equation}

In the reverse process, the DDPM utilizes the iterative sampling from the posterior distribution to get the $x_{t-1}$. This allows the DDPM to generate a sample $x_0 \sim q(x_0)$ from a pure Gaussian noise $x_T \sim \mathcal{N}(0, \mathbf{I})$, where $q(x_0)$ denotes the data distribution of the training dataset.

\section{Method}

In this section, we provide a detailed introduction to our method. We first investigate the properties of \emph{h-space} representation at different time-step in Sec.~\ref{Sec: H-Space}. And then, we introduce the delicately designed DiffLI$^2$D in Sec.~\ref{Sec: DiffLI2R}.

\subsection{H-Space Investigation for Image Dehazing}
\label{Sec: H-Space}

The denoising neural network $\epsilon_{\theta}$ in Eq.~\ref{Eq: Diffusion loss function} is commonly implemented as U-Net in diffusion models. The bottleneck of the frozen pre-trained U-Net, also known as the \emph{h-space}, has been demonstrated to be rich in semantics and can be utilized for high-level semantic manipulation~\cite{kwon2022diffusion}. This inspires us to investigate the potential of \emph{h-space} representation for low-level image dehazing.

To simplify expression, let's define some variables first. Given a hazy image $x$ and its corresponding clean (ground-truth) counterpart $y$, we can get their noisy version at time-step $t$ through the Eq.~\ref{Eq: Diffusion forward 2}, denoted as $x_t$ and $y_t$, respectively, where $t \in [0, T]$. Following~\cite{dhariwal2021diffusion}, the $T$ is set to 1000. Note that $x = x_0$, and $y = y_0$.
By feeding the $x_t$ and $y_t$ into the frozen pre-trained diffusion models $\epsilon_{\theta}$, we can further obtain the corresponding \emph{h-space} features, represented as $h^{haz}_t$ and $h^{cle}_t$, respectively.

\begin{figure*}[t!]
    \centering
    \includegraphics[width=0.98\linewidth]{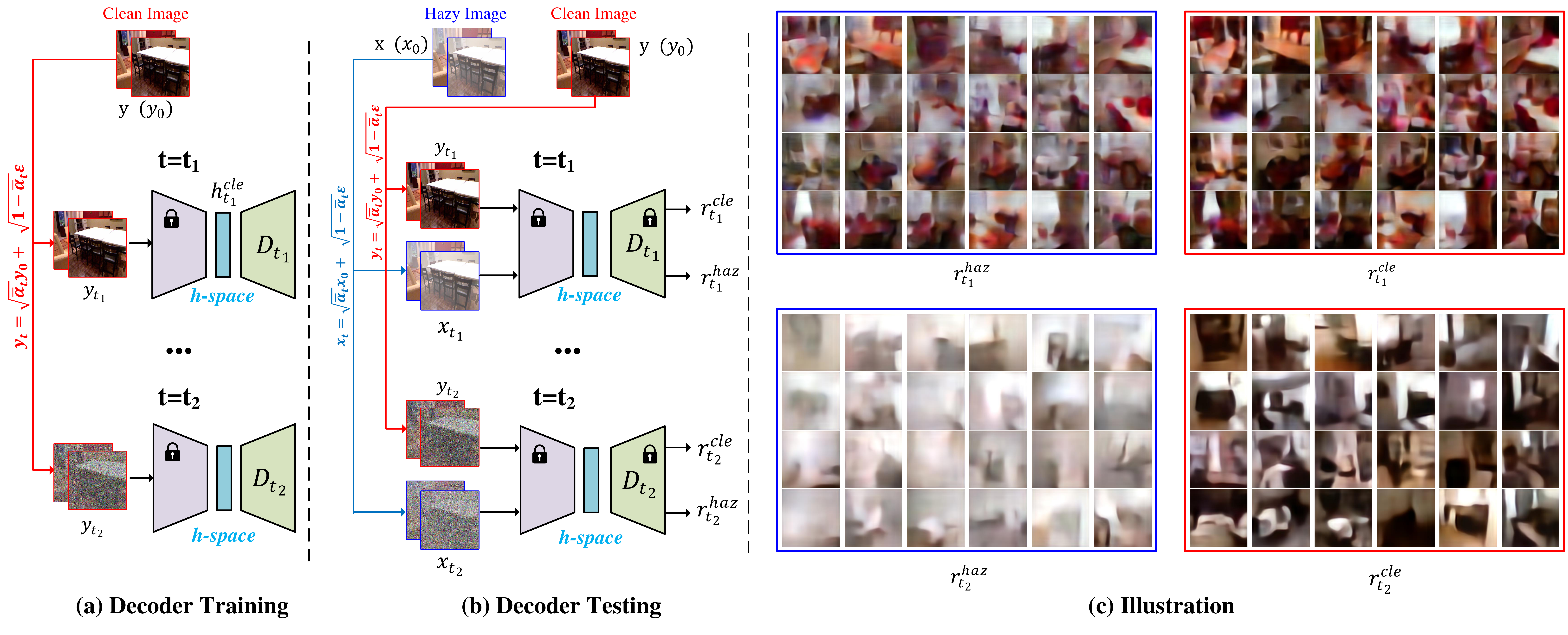}
    \caption{Illustration of our investigation of the \emph{h-space}. (a) Decoder training: for each time-step $t$, we train a decoder $D_t$ to reconstruct the noise-free clean image $y$ from the \emph{h-space} feature $h^{cle}_t$. Note that the decoders are trained with clean noisy images $y_t$ only. (b) Decoder testing: we send both hazy and clean \emph{h-space} features $h^{haz}_t$ and $h^{cle}_t$ to the trained $D_t$, and obtain $r^{haz}_t$ and $r^{cle}_t$, respectively. (c) Illustration of $r^{haz}_t$ and $r^{cle}_t$ at different $t$. As the time-step $t$ changes, the $r^{haz}_t$ represents different components of the original hazy image $x^{haz}$, which further indicates the different representation of \emph{h-space} features at different $t$.}
    \label{Fig: H-Space}
\end{figure*}

\textbf{Investigating H-Space.}
To explore the relation between hazy and clean images in \emph{h-space}, we propose a decoder to map the \emph{h-space} features back to images, as shown in Fig.~\ref{Fig: H-Space}. Specifically, for each time-step $t$, we train a corresponding decoder $D_t$, which aims to map the $h^{cle}_t$ to the noise-free clean image $y$. It can be formulated by:

\begin{equation}
\label{Eq: training for decoder}
    \mathcal{L}_t = \| D_t(h^{cle}_t) - y \|_1
\end{equation}
where $\| \cdot \|_1$ denotes the L1 regularization. Note that we only use the \emph{h-space} features $h^{cle}_t$ corresponding to the clean images $y_t$ to train $D_t$, while the hazy images and their \emph{h-space} features do not participate in training $D_t$.

After that, we feed the $h^{haz}_t$ to the trained $D_t$, and obtain the corresponding reconstruction results, which is $r^{haz}_t = D_t(h^{haz}_t)$. The $h^{cle}_t$ is also sent to the $D_t$ for comparison, which is $r^{cle}_t = D_t(h^{cle}_t)$.
Interestingly, we find that the $r^{haz}_t$ represents different characteristics of the original hazy image $x$, as $t$ changes.
As shown in Fig.~\ref{Fig: H-Space}, when $t$ is small (\eg, $t=t_1$), $r^{haz}_{t_1}$ focuses on representing the contents of the original image, making it very similar to $r^{cle}_{t_1}$. As $t$ progressively increases (\eg, when $t=t_2$), $r^{haz}_{t_2}$ shifts from primarily representing the image content to reflecting the haze characteristics in $x$, which makes it significantly different from $r^{cle}_{t_2}$.
Based on the above observations, we can deduce the conclusion: when the time-step $t$ is small, the \emph{h-space} feature of the hazy image primarily represents the content of the image; as $t$ increases, the \emph{h-space} features shift its emphasis towards representing the haze characteristics of the hazy image.
We provide more analysis and implementation details in \textbf{Appendix}.

To further verify our conclusion, we present the t-SNE maps illustrating the \emph{h-space} features of hazy-clean image pairs at different $t$, as shown in Fig~\ref{Fig: Motivation}. We also show their t-SNE maps in image space for comparison. We can see that, when $t$ is small, the $h^{haz}_t$ and its corresponding clean counterpart $h^{cle}_t$ are tightly clustered together, while the $(h^{haz}_t, h^{cle}_t)$ pairs with different content are separated individually, which indicates that both $h^{haz}_t$ and $h^{cle}_t$ represent the image content. In contrast, when $t$ is large, the \emph{h-space} features of hazy images are clustered together, and those of clean images are clustered together separately, showing that the $h^{haz}_t$ represents the haze characteristics in images.
As a comparison, in the image space, the $x_t$ and $y_t$ are distributed irregularly at all time-step $t$.
It is noteworthy that when $t$ is too large (\eg, $t=T$), both the \emph{h-space} and the image space exhibit irregular patterns. This is because the sufficiently large noise removes both content and haze in the original hazy image.

\textbf{Discussion.}
Many works~\cite{choi2022perception, wang2023exploiting} have proven that diffusion models generate images in a coarse-to-fine manner during the reverse process, and we attribute the properties exhibited by \emph{h-space} to this. At the early steps of the diffusion process (\ie, $t$ is small), the diffusion models focus on fine-grained details, which enables the \emph{h-space} features to fully perceive the background content of the image. In contrast, at the late steps of the diffusion process (\ie, $t$ is large), the diffusion models concentrate more on the coarse attributes, which allows the \emph{h-space} features to represent the foreground haze of the image. More discussion is provided in \textbf{Appendix}.

\begin{figure*}[t!]
    \centering
    \includegraphics[width=1.0\linewidth]{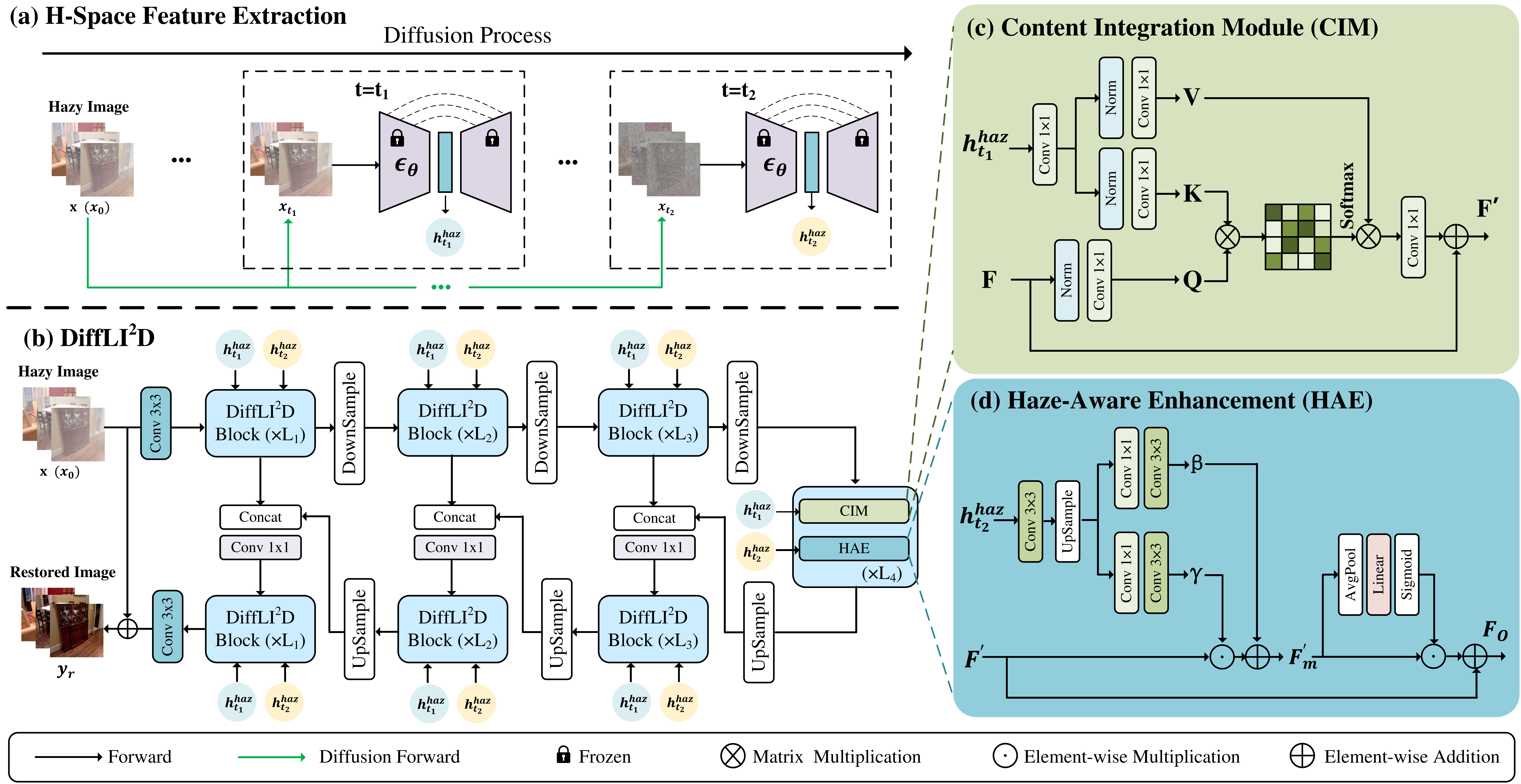}
    \caption{Architecture of the proposed DiffLI$^2$D. (a) Given a hazy image $x$, we first get the noisy versions $x_{t_1}$ and $x_{t_2}$ through Eq.~\ref{Eq: Diffusion forward 2}, and then obtain the \emph{h-space} features $h^{haz}_{t_1}$ and $h^{haz}_{t_2}$ by sending $x_{t_1}$ and $x_{t_2}$ into the frozen pre-trained diffusion model. (b) Architecture of the dehazing network, which comprises multiple blocks arranged in the U-Net structure. Each block consists of a Content Integration Module (CIM) and a Haze-Aware Enhancement (HAE) module, where the former is designed to leverage the $h^{haz}_{t_1}$ to facilitate the content recovery, while the latter utilizes the $h^{haz}_{t_2}$ as guidance for effective haze removal.}
    \label{Fig: DiffLI2R}
\end{figure*}

\subsection{Architecture of the DiffLI$^2$D}
\label{Sec: DiffLI2R}

The observations in Sec.~\ref{Sec: H-Space} promote us to utilize the \emph{h-space} features to facilitate image dehazing. For all \emph{h-space} features $h^{haz}_t$ $(t \in [0, T])$, the $h^{haz}_{t_1}$ (\ie, $t=t_1$) and $h^{haz}_{t_2}$ (\ie, $t=t_2$) are the most representative, where the former captures the underlying content of $x$, while the latter characterizes the haze attributes, as discussed in Sec.~\ref{Sec: H-Space}. The selection of $t_1$ and $t_2$ will be described in Sec.~\ref{Sec: ablation study}.
Based on this, we propose a Diffusion Latent Inspired network for Image Dehazing (DiffLI$^2$D). As shown in Fig.~\ref{Fig: DiffLI2R}, the DiffLI$^2$D comprises multiple DiffLI$^2$D blocks arranged in the U-Net structure. Each block consists of a Content Integration Module (CIM) that leverages $h^{haz}_{t_1}$ for image content recovery, and a Haze-Aware Enhancement (HAE) module which utilizes $h^{haz}_{t_2}$ for haze removal.

\textbf{Content Integration Module.} 
To effectively leverage the \emph{h-space} feature $h^{haz}_{t_1}$ for content recovery, we propose the Content Integration Module (CIM), as illustrated in Fig.~\ref{Fig: DiffLI2R}(c). Specifically, given the intermediate feature $F \in \mathbb{R}^{\hat{H} \times \hat{W} \times \hat{C}}$ and the \emph{h-space} feature $h^{haz}_{t_1} \in \mathbb{R}^{H' \times W' \times C'}$, we can get $h^{haz'}_{t_1} = W_c h^{haz}_{t_1}$, where $h^{haz'}_{t_1} \in \mathbb{R}^{H' \times W' \times \hat{C}}$, and $W_c$ is a $1 \times 1$ convolution kernel. Then, $F$ is projected into query $Q = W_Q F$, and $h^{haz'}_{t_1}$ is projected into key $K = W_K h^{haz'}_{t_1}$ and value $V = W_V h^{haz'}_{t_1}$. $W_Q$, $W_K$, and $W_V$ are all implemented by $1 \times 1$ convolution kernel. After that, we reshape $Q \in \mathbb{R}^{\hat{H} \hat{W} \times \hat{C}}$, and $K, V \in \mathbb{R}^{H'W' \times \hat{C}}$, and obtain final output $F'$ formulated by:

\begin{equation}
\label{Eq: CIM Attention Map}
    F' = W_O \cdot \textit{Softmax}(\frac{Q K^{T}}{\sqrt{\hat{C}}}) \cdot V + F,
\end{equation}
where $W_O$ is implemented by $1\times 1$ convolution kernel. Similar with transformer~\cite{vaswani2017attention}, the multi-head mechanism is introduced to the CIM. Through the interaction, the CIM encourages the DiffLI$^2$D to fully explore the correspondence between $F$ and $h^{haz}_{t_1}$, and further enables the DiffLI$^2$D to dynamically capture the informative content representations in $h^{haz}_{t_1}$ for effective content recovery.

\textbf{Haze-Aware Enhancement.}
We further design a Haze-Aware Enhancement (HAE) module to utilize $h^{haz}_{t_2}$ as guidance for haze removal, as shown in Fig.~\ref{Fig: DiffLI2R}(d). Given the \emph{h-space} feature $h^{haz}_{t_2}$, we first use it as the guidance to dynamically enhance the input feature $F'$ in a SFT~\cite{wang2018recovering} manner, which is:

\begin{equation}
\label{Eq: DGE gamma and beta}
   F'_m = W_{\gamma} h^{haz}_{t_2} \cdot F' + W_{\beta} h^{haz}_{t_2}.
\end{equation}

After that, we further modulate the integrated feature $F'_m$ from channel dimension, which is:

\begin{equation}
    F_o = \sigma(W_L \cdot \textit{AvgPool}(F'_m)) \cdot F'_m + F',
\end{equation}
where $\sigma$ denotes the Sigmoid function, $W_L$ is a linear layer, and $\textit{AvgPool}$ means the average pooling operation. Through this, the HAE adapts DiffLI$^2$D to the haze characteristics within $h^{haz}_{t_2}$, which dynamically modulates the input features under the guidance of $h^{haz}_{t_2}$ for haze removal and further enhancement.

The optimization objective for training DiffLI$^2$D is defined as:

\begin{equation}
\label{Eq: loss}
    \mathcal{L} = \| y_r - y \|_1
\end{equation}
where $y_r$ denotes the restored image. Note that the pre-trained diffusion model $\epsilon_{\theta}$ is frozen, and do not participate in the optimization.

Note that our DiffLI$^2$D avoids re-training diffusion models~\cite{yue2023resshift, jiang2023low} and the time-consuming reverse sampling process~\cite{chung2022diffusion, song2022pseudoinverse}. Instead, by leveraging the \emph{h-space} features $h^{haz}_{t_1}$ and $h^{haz}_{t_2}$ as guidance, the DiffLI$^2$D can effectively recover clean images. More detailed comparisons are discussed in \textbf{Appendix}.

\section{Experiments}
\label{Sec: Experiments}

\makeatletter
\newcommand\semismall{\@setfontsize\semismall{8}{10.5}}
\makeatother

\begin{table*}[t!]
\semismall
\centering
\caption{Performance comparisons with state-of-the-art dehazing methods across synthetic (\ie, SOTS) and real-world (\ie Dense-Haze and NH-HAZE) dehazing datasets. The superscript $^*$ means diffusion model-based method for image dehazing.  The best results are marked as \textbf{bold} and the second ones are masked by \underline{underline}.}
\begin{tabular}{c|ccc|ccc|ccc|c}
    \shline
    \centering
    \multirow{2}{*}{Method} & \multicolumn{3}{c|}{SOTS} & \multicolumn{3}{c|}{Dense-Haze} & \multicolumn{3}{c|}{NH-HAZE} & \multirow{2}{*}{\#Params} \\
    \cline{2-10}
    \multicolumn{1}{c|}{} & PSNR & SSIM & LPIPS & PSNR & SSIM & LPIPS & PSNR & SSIM & LPIPS & \multicolumn{1}{c}{} \\ 
    \hline 

    DCP~\cite{he2010single} & 15.09 & 0.765 & 0.069 & 10.03 & 0.386 & 0.605 & 10.57 & 0.520 & 0.399 & - \\
    DehazeNet~\cite{cai2016dehazenet} & 20.64 & 0.800 & 0.242 & 13.84 & 0.425 & 0.637 & 16.62 & 0.524 & 0.529 & 0.01M \\
    AOD-Net~\cite{li2017aod} & 19.82 & 0.818 & 0.099 & 13.14 & 0.414 & 0.599 & 15.40 & 0.569 & 0.495 & 0.002M \\
    FFA-Net~\cite{qin2020ffa} & 36.39 & 0.989 & \underline{0.005} & 14.39 & 0.452 & 0.498 & 19.87 & 0.692 & 0.365 & 4.68M \\
    MSBDN~\cite{dong2020multi} & 33.79 & 0.984 & 0.029 & 15.37 & 0.486 & 0.536 & 19.23 & \underline{0.706} & 0.292 & 31.35M \\
    SwinIR~\cite{liang2021swinir} & 24.93 & 0.932 & 0.049 & 12.20 & 0.510 & 0.639 & 16.15 & 0.623 & 0.479 & 0.91M \\
    AECR-Net~\cite{wu2021contrastive} & 37.17 & \underline{0.990} & 0.007 & 15.80 & 0.466 & 0.537 & 19.88 & 0.707 & 0.278 & 2.61M \\
    MPRNet~\cite{zamir2021multi} & 32.14 & 0.983 & 0.011 & 13.82 & 0.519 & 0.620 & 17.88 & 0.631 & 0.368 & 15.74M \\
    Restormer~\cite{zamir2022restormer} & \underline{38.43} & 0.989 & 0.009 & 15.17 & 0.557 & 0.629 & 18.32 & 0.635 & 0.355 & 26.13M \\
    Dehamer~\cite{guo2022image} & 36.63 & 0.988 & \underline{0.005} & \underline{16.62} & \underline{0.560} & \underline{0.480} & \textbf{20.66} & 0.684 & \underline{0.230} & 132.50M \\
    IR-SDE$^*$~\cite{luo2023image} & 33.82 & 0.984 & 0.014 & 12.03 & 0.508 & 0.485 & 12.59 & 0.520 & 0.361 & 537.21M \\
    ResShift$^*$~\cite{yue2023resshift} & 29.06 & 0.950 & 0.017 & 13.67 & 0.517 & 0.576 & 16.26 & 0.625 & 0.327 & 114.65M \\
    \hline
    DiffLI$^2$D$^*$ (Ours) & \textbf{40.33} & \textbf{0.992} & \textbf{0.004} & \textbf{16.97} & \textbf{0.584} & \textbf{0.406} & \underline{20.29} & \textbf{0.738} & \textbf{0.217} & 8.63M \\
    
    \shline
\end{tabular}\\

\label{Tab: Dehazing}
\end{table*}

\begin{figure*}[h]
    \centering
    \includegraphics[width=1.0\textwidth]{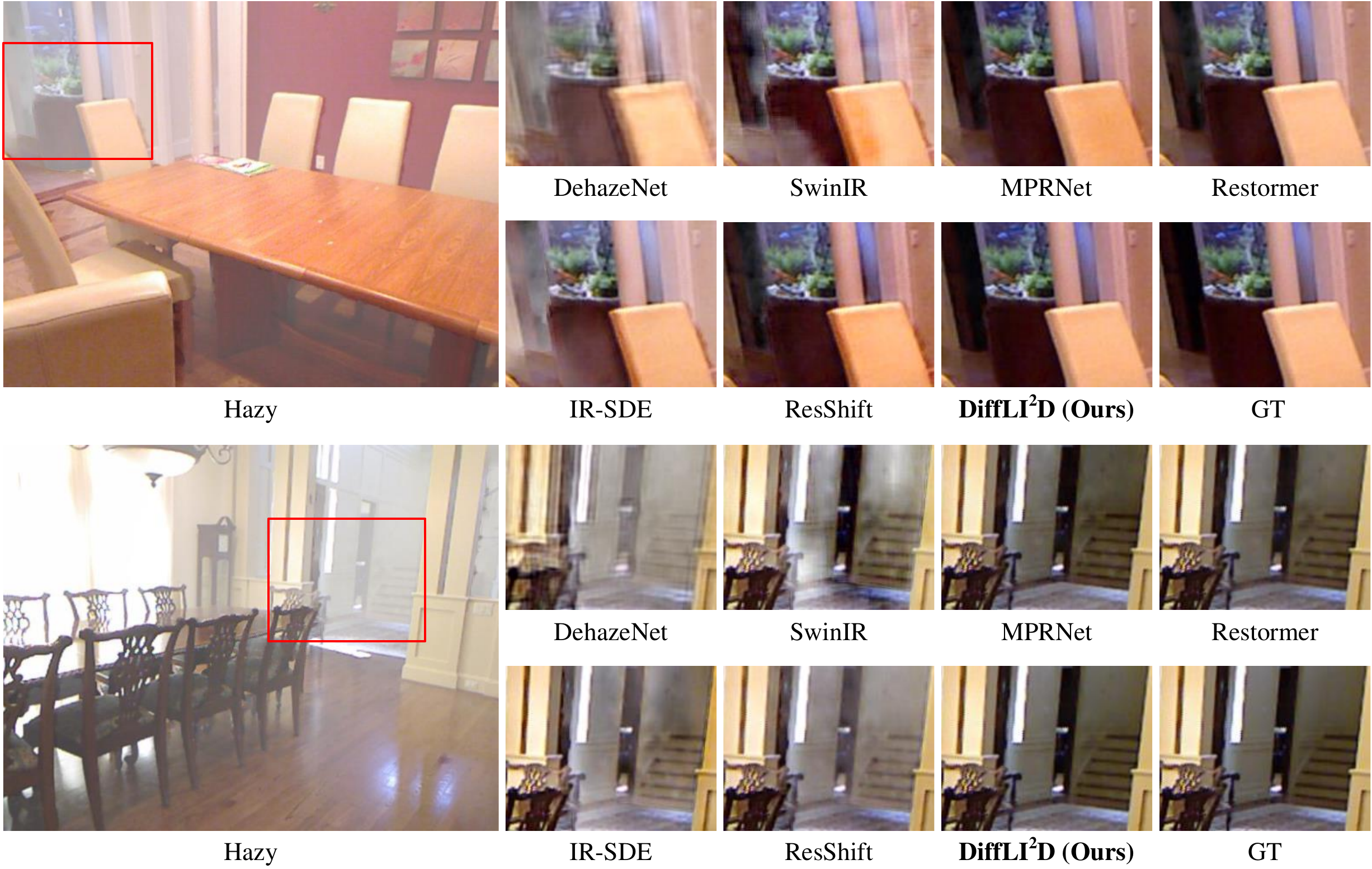}
    \caption{Qualitative results of different methods for dehazing on SOTS dataset. Our method is shown in \textbf{bold}.}
    \label{Fig: ITS}
\end{figure*}

In this section, we conduct comprehensive experiments to verify the effectiveness of the proposed DiffLI$^2$D. More experimental results and details can be found in \textbf{Appendix} for further reference.

\subsection{Implementation Details}

\textbf{Datasets.}
We evaluate our method on both synthetic and real-world datasets. For synthetic scene, the RESIDE~\cite{li2018benchmarking} is utilized for training and testing. Specifically, the subset Indoor Training Set (ITS) of RESIDE is used for training. It consists of 13,990 hazy images, which are generated from 1,399 clean images. The subset Synthetic Objective Testing Set (SOTS) of RESIDE includes 500 indoor and 500 outdoor hazy images. We choose the indoor part for testing. For real-world scene, the Dense-Haze~\cite{ancuti2019dense} and NH-HAZE~\cite{ancuti2020nh} are adopted. Both datasets have 55 hazy-clean image pairs, 50 of which are utilized for training and 5 of which are utilized for testing.

\textbf{Training Details.}
The proposed DiffLI$^2$D is trained by Adam optimizer, where $\beta_1$ and $\beta_2$ are set to 0.9 and 0.999, respectively. The total training epoch is set to 1200. The initial learning rate is set to $2 \times 10^{-4}$, and decreases with a factor of 0.5 every 300 epochs. The mini-batch is set to 40, and the images are resized, cropped to $128 \times 128$ with being flipped horizontally randomly. In our experiments, we choose the unconditional DDPM model pre-trained on ImageNet~\cite{dhariwal2021diffusion} as the $\epsilon_{\theta}$. The whole model is trained with one 3090Ti GPU using PyTorch framework. More implementation details are provided in \textbf{Appendix}.

\textbf{Evaluation Metrics.}
To assess the performance, we adopt three different metrics, including PSNR, SSIM~\cite{wang2004image} and LPIPS~\cite{zhang2018unreasonable}. PSNR and SSIM are employed to quantify the fidelity of the restored images -- the higher their values, the better the restoration quality. LPIPS is utilized to measure the perceptual difference and visual quality, with lower values indicating better performance.

\subsection{Evaluation on Image Dehazing}

\textbf{Experiment Results on Synthetic Dataset.}
Tab.~\ref{Tab: Dehazing} shows the comparison results between the DiffLI$^2$D and existing dehazing methods on SOTS dataset. We can see that DiffLI$^2$D outperforms Restormer and Dehamer with less parameters. Additional, the lower LPIPS scores indicate that the image restored by DiffLI$^2$D are better aligned with human visual system. It is noteworthy that, compared with existing diffusion model-based methods (\eg, IR-SDE), our method not only achieves superior results but also avoids the re-training diffusion models. Moreover, our method circumvents the potential time-consuming sampling during the inference. We also shows the qualitative comparison in Fig.~\ref{Fig: ITS}. As we can see, our method can recover image details and remove haze more effectively.

\begin{figure*}[t!]
    \centering
    \includegraphics[width=1.0\textwidth]{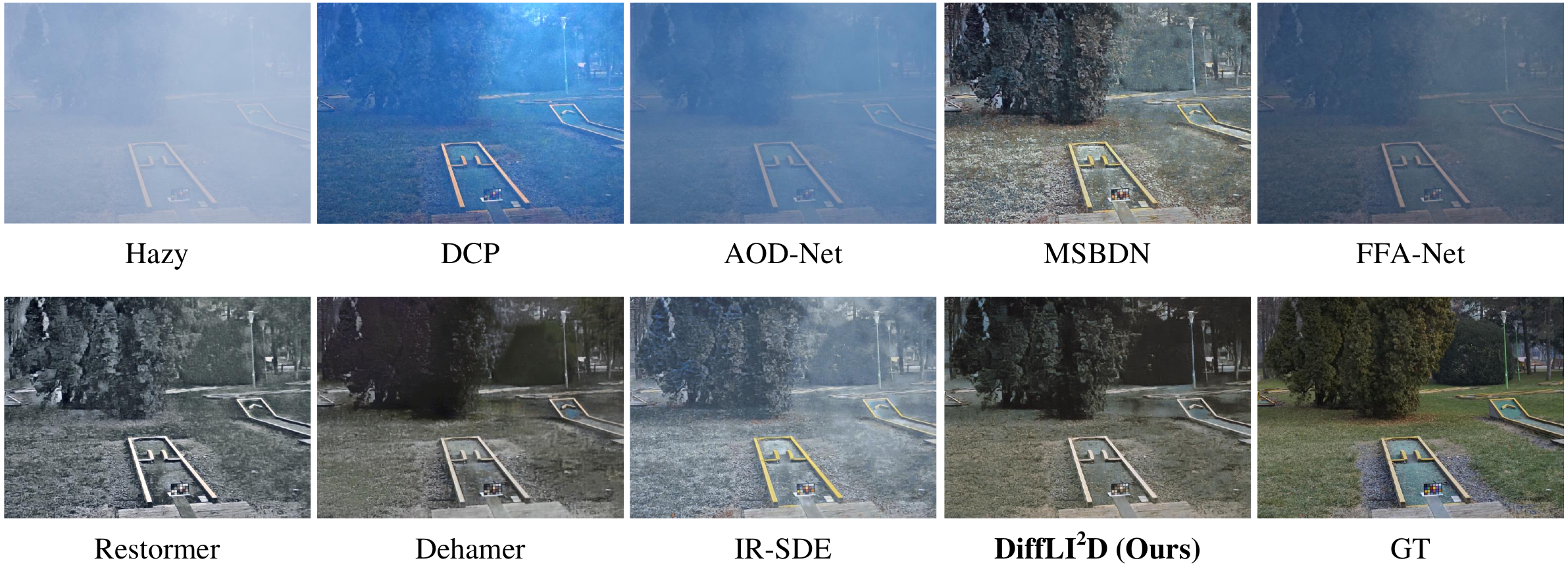}
    \caption{Qualitative results of different methods for dehazing on Dense-Haze dataset.}
    \label{Fig: Dense-Haze}
\end{figure*}

\begin{figure*}[t!]
    \centering
    \includegraphics[width=1.0\textwidth]{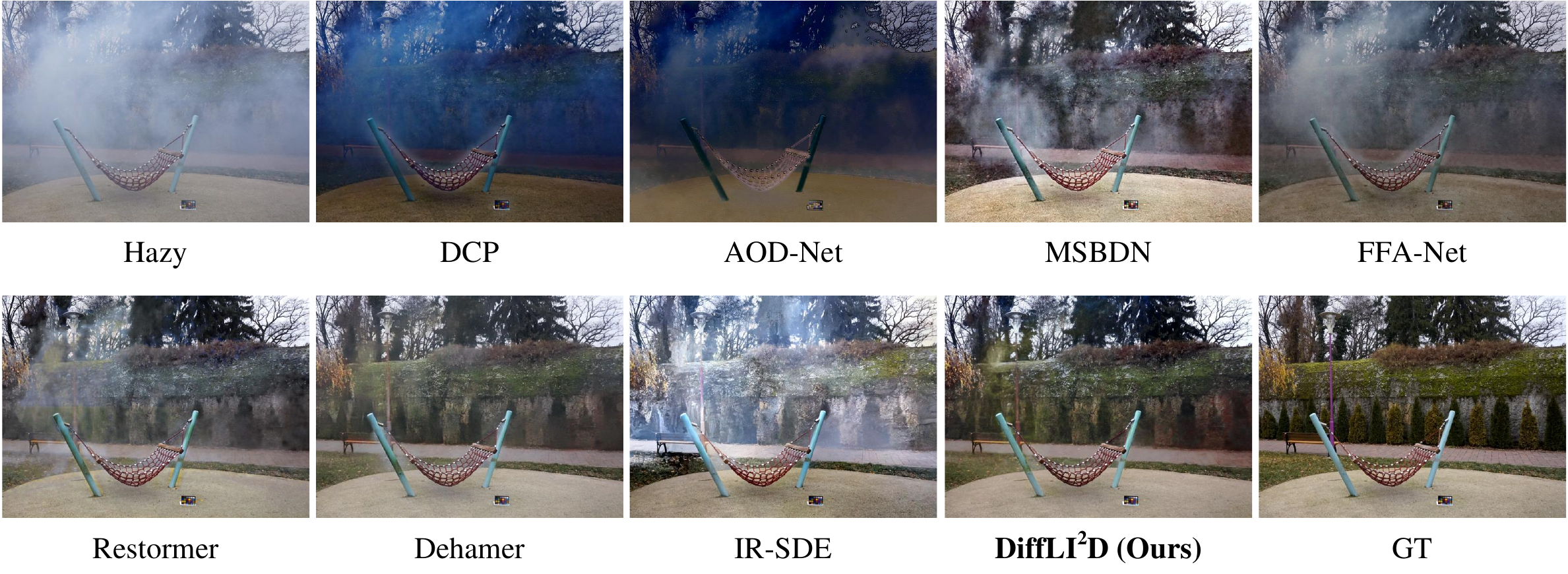}
    \caption{Qualitative results of different methods for dehazing on NH-HAZE dataset.}
    \label{Fig: NH-HAZE}
\end{figure*}

\textbf{Experiment Results on Real-World Datasets.}
We further evaluate the proposed DiffLI$^2$D on Dense-Haze~\cite{ancuti2019dense} and NH-HAZE~\cite{ancuti2020nh} dataset. Tab.~\ref{Tab: Dehazing} shows that the DiffLI$^2$D outperforms or achieves at least comparable performance to compared image dehazing methods across two datasets. Fig.~\ref{Fig: Dense-Haze} and Fig.~\ref{Fig: NH-HAZE} illustrate the qualitative results on Dense-Haze and NH-HAZE, respectively. It can be seen that our method can recover haze-free images with better visual effects.

\subsection{Ablation Study}
\label{Sec: ablation study}

In this section, we perform comprehensive ablation studies to demonstrate the effectiveness of our designs in the proposed DiffLI$^2$D. More results of ablation studies are provided in \textbf{Appendix}.

\begin{figure}[t!]
    \centering
    \includegraphics[width=1.0\linewidth]{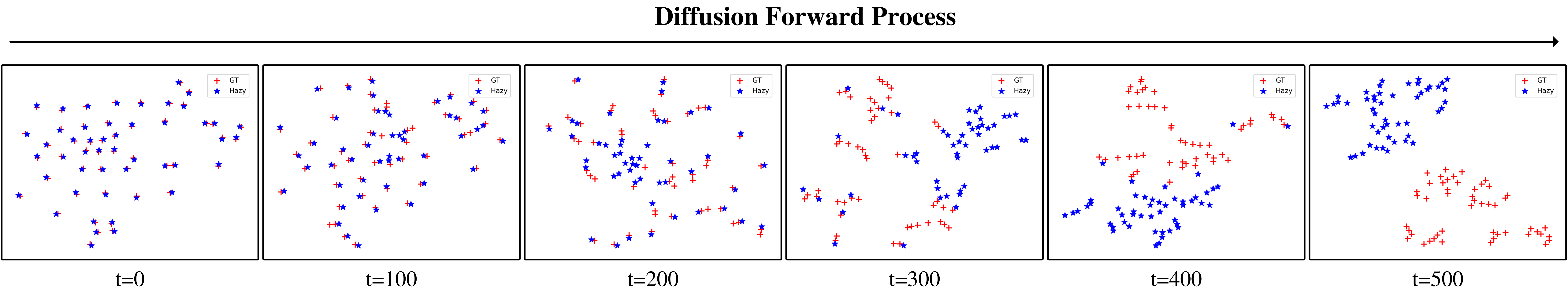}
    \caption{Transformations of the \emph{h-space} feature distributions of hazy-clean image pairs, as the time-step $t$ changes.}
    \label{Fig: transformation}
\end{figure}

\begin{figure}[t!]
    \centering
    \includegraphics[width=1.0\linewidth]{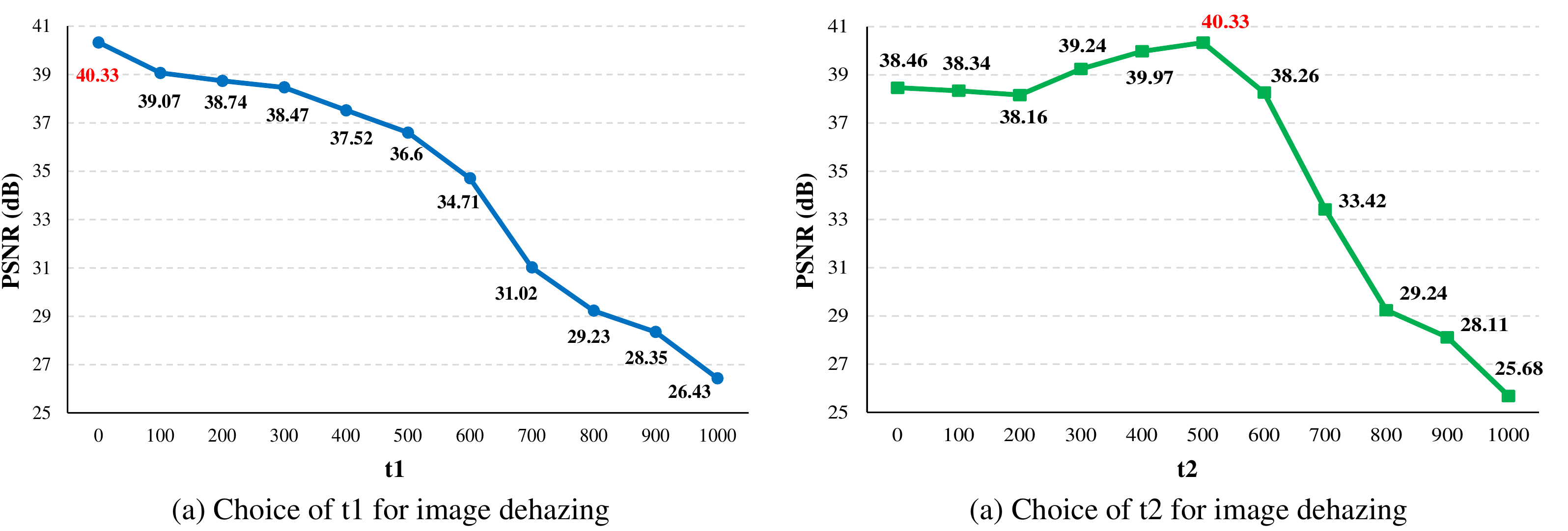}
    \caption{Performance (PSNR) comparison of different choice of $t_1$ and $t_2$ for image dehazing on SOTS dataset.}
    \label{Fig: Choice of t1 and t2}
\end{figure}

\textbf{Choice of $t_1$ and $t_2$ for image dehazing.}
As discussed in Sec.~\ref{Sec: H-Space}, the $\emph{h-space}$ features exhibit different characteristics at different time-step $t$. In fact, as $t$ increases, the \emph{h-space} features undergo a gradual and continuous transformation, transitioning from primarily representing the contents to mainly reflecting the haze properties. We show this transformation in Fig.~\ref{Fig: transformation}. This implies that, compared with other time-steps, the \emph{h-space} feature $h^{haz}_0$ corresponding to $t=0$ is the most representative of the underlying image content. So in our experiments, $t_1$ is set to $0$. As illustrated in Fig.~\ref{Fig: Choice of t1 and t2}(a), $t_1 = 0$ is the best choice.

We further investigate the influence of the choice of $t_2$ for image dehazing. We find that the DiffLI$^2$D achieve optimal results for image dehazing when $t_2$ is around 500. As described in Fig.~\ref{Fig: Choice of t1 and t2}(b), $t_2 = 500$ outperforms $t_2 = 100$ by 1.99dB in terms of PSNR, which also surpasses $t_2 = 600$ by 2.07dB in PSNR. This is consistent with the observation in Fig.~\ref{Fig: transformation}. When $t_2$ is relatively small (\eg, $t_2 = 300$), the \emph{h-space} features $h^{haz}_{t_2}$ are still intertwined with $h^{cle}_{t_2}$, and cannot effectively represent the haze characteristics. When $t_2$ is around 500, the $h^{haz}_{t_2}$ are clustered together, which can capture and represent the haze attributes effectively.
Note that whether for $t_1$ or $t_2$, when they are too large (\eg, $t_1=1000$ or $t_2=1000$), the image dehazing performance of DiffLI$^2$D experiences a significant decline. This can be attributed to the excessive noise that erases a considerable amount of information from the original hazy image $x$, making the \emph{h-space} feature less effective in guiding the dehazing process.

\textbf{Effectiveness of the CIM and HAE.}
Our DiffLI$^2$D consists of two key modules, the content integration module (CIM) and the haze-aware enhancement (HAE) module. The former is designed to facilitate the image content recovery with $h^{haz}_{t_1}$, while the latter aims to enhance haze removal using $h^{haz}_{t_2}$. To evaluate the benefits of them, we design several variants as shown in Tab.~\ref{Tab: Effectiveness of the CIM and DGE}. Among them, the ``DiffLI$^2$D w/o CIM'' represents that we replace the CIM in each DiffLI$^2$D block with self-attention module that has similar number of parameter as CIM. We do a similar operation for the ``DiffLI$^2$D w/o HAE''. The ``Baseline'' means that both CIM and HAE are replaced.

As we can see, ``DiffLI$^2$D w/o CIM'' and ``DiffLI$^2$D w/o HAE'' outperform ``Baseline'' by 1.56dB, 1.72dB, 0.53dB and 1.79dB, 1.83dB, 1.05dB in terms of PSNR on SOTS, Dense-Haze, NH-HAZE dataset, respectively. With both two modules, the ``DiffLI$^2$D'' achieves 40.33dB, 16.97dB, 20.29dB on SOTS, Dense-Haze, NH-HAZE dataset, which demonstrates that the CIM and HAE are complementary and both vital to the DiffLI$^2$D, jointly resulting in a superior performance in image dehazing.

\begin{table}[t!]
\caption{Ablation results of several variants of DiffLI$^2$D for image dehazing on SOTS, Dense-Haze, and NH-HAZE dataset. The PSNR is utilized for evaluation.}
\small
    \centering
    \begin{tabular}{C{3.3cm}|C{0.9cm}C{0.9cm}|C{1.4cm}C{1.7cm}C{1.7cm}}
        \shline
         Model & CIM & HAE & SOTS & Dense-Haze & NH-HAZE \\
         \hline
         Baseline & $\times$ & $\times$ & 36.65 & 14.32 & 18.11 \\ 
         DiffLI$^2$D w/o CIM & $\times$ & $\surd$ & 38.21 & 16.04 & 18.64 \\ 
         DiffLI$^2$D w/o HAE & $\surd$ & $\times$ & 38.44 & 16.15 & 19.16  \\ 
         DiffLI$^2$D & $\surd$ & $\surd$ & \textbf{40.33} & \textbf{16.97} & \textbf{20.29} \\
         \shline
    \end{tabular}
    \label{Tab: Effectiveness of the CIM and DGE}
\end{table}

\begin{table*}[t!]
\small
\centering
\caption{Performance (PSNR) comparisons between \emph{H-Space} and other feature spaces for image dehazing on SOTS, Dense-Haze, and NH-HAZE datasets.}
\begin{tabular}{C{2.5cm}|C{1.4cm}C{1.4cm}C{1.4cm}|C{1.4cm}C{1.4cm}C{1.4cm}}
    \shline
    \centering
    \multirow{2}{*}{Method} & \multicolumn{3}{c|}{Layer Comparison} & \multicolumn{3}{c}{Network Comparison}  \\
    \cline{2-7}
    \multicolumn{1}{c|}{} & Layer-E & Layer-D & H-Space & VGG16 & ResNet50 & H-Space  \\ 
    \hline 
    SOTS & 36.52 & 39.10 & \textbf{40.33} & 27.83 & 25.49 & \textbf{40.33} \\
    Dense-Haze & 13.86 & 16.44 & \textbf{16.97} & 12.87 & 12.52 & \textbf{16.97} \\
    NH-HAZE & 17.53 & 19.88 & \textbf{20.29} & 16.23 & 15.62 & \textbf{20.29} \\
    \shline
\end{tabular}\\

\label{Tab: Different Spaces}
\end{table*}

\textbf{Effectiveness of the \emph{H-Space} features.}
To further evaluate the effectiveness of the \emph{h-space} for image dehazing, we compare it with other feature spaces. Specifically, we compare \emph{h-space} with feature spaces of different layers in the same diffusion model, and we also compare \emph{h-space} with feature spaces of other networks. Tab.~\ref{Tab: Different Spaces} shows the comparison results. For different layers, ``Layer-E'' and ``Layer-D'' means that we replace the \emph{h-space} representations $h^{haz}_{t_1}$ and $h^{haz}_{t_2}$ with those features extracted from the encoder and decoder of the same pre-trained diffusion model. As we can see, compared with features extracted from the encoder and decoder, the \emph{h-space} features achieves 3.81dB, 3.11dB, 2.76dB and 1.23dB, 0.53dB, 0.41dB improvement in terms of PSNR on SOTS, Dense-Haze, and NH-HAZE dataset, respectively. For different networks, we choose the commonly-used VGG16~\cite{simonyan2014very} and ResNet50~\cite{he2016deep} as comparison. Note that, for fair comparison, both VGG16 and ResNet50 are pre-trained on ImageNet~\cite{russakovsky2015imagenet}, which is the same dataset used to pre-train the diffusion model $\epsilon_{\theta}$ employed in our study. The ``VGG16'' and ``ResNet50'' in Tab.~\ref{Tab: Different Spaces} denote that we replace the \emph{h-space} features $h^{haz}_{t_1}$ and $h^{haz}_{t_2}$ with  corresponding features extracted from the feature spaces of VGG16 and ResNet50, respectively. As we can see, \emph{h-space} features drive from pre-trained diffusion model can facilitate the image dehazing more effectively than those extracted from other pre-trained models.

\subsection{The Generalization Capabilities}
\begin{figure}[t!]
    \centering
    \includegraphics[width=1.0\linewidth]{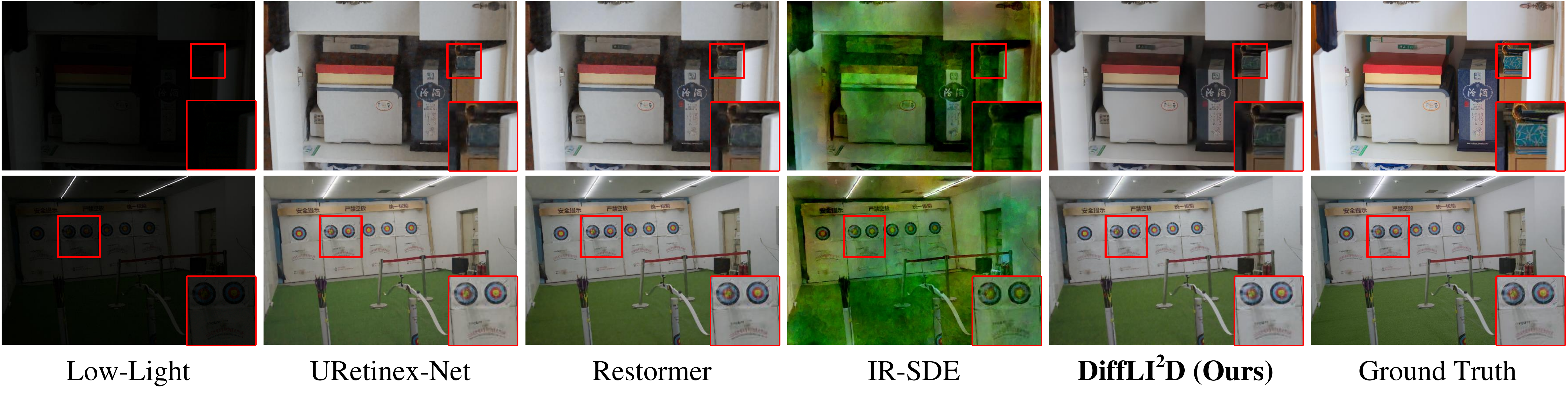}
    \caption{Qualitative results of different methods for low-light image enhancement.}
    \label{Fig: Enhancement}
\end{figure}

To evaluate the generalization capabilities of our method, we further evaluate the effectiveness of our method on more low-level image restoration tasks. Specifically, we further test our method on low-light image enhancement on LOL-v1~\cite{wei2018deep} and LOL-v2 dataset~\cite{yang2021sparse}. We show the comparison results about low-light image enhancement on Tab.~\ref{Tab: Comparison on Enhancement}. FID~\cite{heusel2017gans} metric is introduced for further evaluation. We also show the qualitative results in Fig.~\ref{Fig: Enhancement}.
As we can see, the proposed DiffLI$^2$D can also achieve superior performance when handling other low-level image restoration tasks, which demonstrates that the capabilities of DiffLI$^2$D are not limited to image dehazing. More discussion, analysis, visual results and comparison results about other image restoration tasks are provided in \textbf{Appendix}.

\begin{table*}[t!]
\semismall
\centering
\caption{Performance comparisons with state-of-the-art low-light image enhancement methods on LOL-v1 and LOL-v2 Real dataset. The superscript $^*$ denotes the diffusion model-based image restoration method. The best results are marked as \textbf{bold} and the second ones are masked by \underline{underline}.}
\begin{tabular}{C{2.6cm}|C{1.0cm}C{1.0cm}C{1.1cm}C{1.0cm}|C{1.0cm}C{1.0cm}C{1.1cm}C{1.0cm}}
    \shline
    \centering
    \multirow{2}{*}{Method} & \multicolumn{4}{c|}{LOLv1} & \multicolumn{4}{c}{LOLv2-Real} \\
    \cline{2-9}
    \multicolumn{1}{c|}{} & PSNR & SSIM & LPIPS & FID & PSNR & SSIM & LPIPS  & FID \\ 
    \hline 
    RetinexNet~\cite{wei2018deep} & 16.77 & 0.462 & 0.417 & 126.27 & 17.72 & 0.652 & 0.436 & 133.91 \\
    Zero-DCE~\cite{guo2020zero} & 14.86 & 0.562 & 0.372 & 87.24 & 18.06 & 0.580 & 0.352 & 80.45 \\
    EnlightenGAN~\cite{jiang2021enlightengan} & 17.61 & 0.653 & 0.372 & 94.70 & 18.68 & 0.678 & 0.364 & 84.04 \\
    URetinex-Net~\cite{wu2022uretinex} & 19.97 & \underline{0.828} & \underline{0.267} & 62.38 & 21.13 & 0.827 & \underline{0.208} & \underline{49.84} \\
    Uformer~\cite{wang2022uformer} & 19.00 & 0.741 & 0.354 & 109.35 & 18.44 & 0.759 & 0.347 & 98.14 \\
    Restormer~\cite{zamir2022restormer} & \underline{20.61} & 0.792 & 0.288 & 73.00 & \underline{21.36} & \underline{0.835} & 0.314 & 63.69 \\
    WeatherDiff$^*$~\cite{ozdenizci2023restoring} & 16.30 & 0.786 & 0.277 & 65.61 & 15.87 & 0.801 & 0.272 & 65.82 \\
    ResShift$^*$~\cite{yue2023resshift} & 19.23 & 0.735 & 0.225 & \underline{61.21} & 20.41 & 0.704 & 0.218 & 60.72 \\
    IR-SDE$^*$~\cite{luo2023image} & 12.90 & 0.557 & 0.486 & 175.33 & 12.53 & 0.511 & 0.453 & 157.08 \\
    GDP$^*$~\cite{fei2023generative} & 13.93 & 0.630 & 0.445 & 95.16 & 13.15 & 0.527 & 0.421 & 97.54 \\
    DiffLI$^2$D$^*$ (Ours) & \textbf{23.30} & \textbf{0.849} & \textbf{0.136} & \textbf{55.88} & \textbf{22.35} & \textbf{0.874} & \textbf{0.186} & \textbf{42.49} \\
    \shline
\end{tabular}\\
\label{Tab: Comparison on Enhancement}
\end{table*}

\section{Conclusion}

In this paper, we investigate the semantic latent space of frozen pre-trained diffusion models for image dehazing, and reveal that the features in the semantic latent space can effectively represent the content and haze characteristics of hazy images, as the time-step changes. We also propose a Diffusion Latent Inspired network for Image Dehazing (DiffLI$^2$D), which uses the semantic latent features of frozen pre-trained diffusion models for effective image dehazing. Extensive experiments on multiple datasets demonstrate the effectiveness of our method. \\

\noindent\textbf{Acknowledgments.} This work was supported by the Anhui Provincial Natural Science Foundation under Grant 2108085UD12. We acknowledge the support of GPU cluster built by MCC Lab of Information Science and Technology Institution, USTC.

\bibliographystyle{splncs04}
\bibliography{main}

\newpage
\appendix

\noindent\textbf{\Large Appendix}

\section{More Analysis and Discussion on H-Space}

In the Sec.4.1 of the main body, we have demonstrated that the \emph{h-space} features can effectively represent the underlying content and haze characteristics of the hazy image, as the diffusion time-step $t$ changes. In this section, we provide more analysis and discussion about the \emph{h-space}.

\subsection{More Analysis about the H-Space}

In Sec.4.1 of the main body, we investigate the \emph{h-space} features by training a decoder $D_t$ for each $t$. Since the training objective for decoder $D_t$ is to map the \emph{h-space} feature $h^{cle}_t$ corresponding to $y_t$ back to its original noise-free version $y$, and considering that only clean images are utilized for training, the $D_t$ can fully perceive how the \emph{h-space} feature $h^{cle}_t$ represents the original clean image $y$ at the time-step $t$. Note that $y_t$ is the noisy version of $y$ at the time-step $t$.

And then, we send both the \emph{h-space} features of hazy images $h^{haz}_t$ and the \emph{h-space} features of clean images $h^{cle}_t$ to the trained decoder $D_t$, and get the corresponding reconstruction results $r^{haz}_t$ and $r^{cle}_t$, respectively. Since the $D_t$ is trained on clean images, it is not surprising that the $r^{cle}_t$ can represent the content of the original clean image $y$ across all time-step $t$. However, the $r^{haz}_t$ varies significantly for different time-step $t$. As shown in Fig.2(c) of the main body, when $t$ is small, the $r^{haz}_t$ is similar with $r^{cle}_t$, both representing the content in the original image. This means that, when $t$ is small, both $h^{haz}_t$ and $h^{cle}_t$ encode the content in the original image. In contrast, when $t$ is large, one can barely see underlying content of $x_t$ from $r^{haz}_t$, instead see the haze characteristics in $x_t$. This is a very interesting and meaningful phenomenon. By precisely manipulating the noise added to the hazy image $x$, we can obtain the \emph{h-space} features that represent different components of $x$. And this interesting phenomenon leads us to the conclusion presented in Sec.4.1 of the main body.

\begin{figure}[t!]
    \centering
    \includegraphics[width=0.92\linewidth]{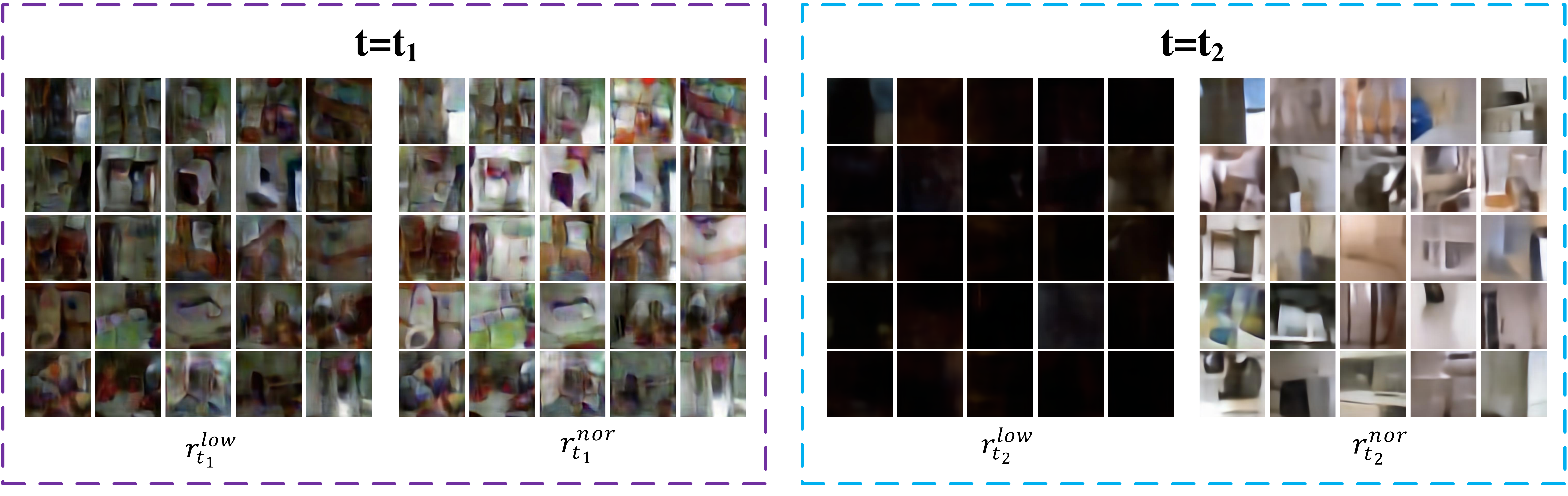}
    \caption{Illustration of $r^{low}_t$ and $r^{nor}_t$ at different time-step $t$ on low-light/normal-light image pairs.}
    \label{Fig: H-Space on Low-Light}
\end{figure}

\begin{figure*}[t!]
    \centering
    \includegraphics[width=0.92\linewidth]{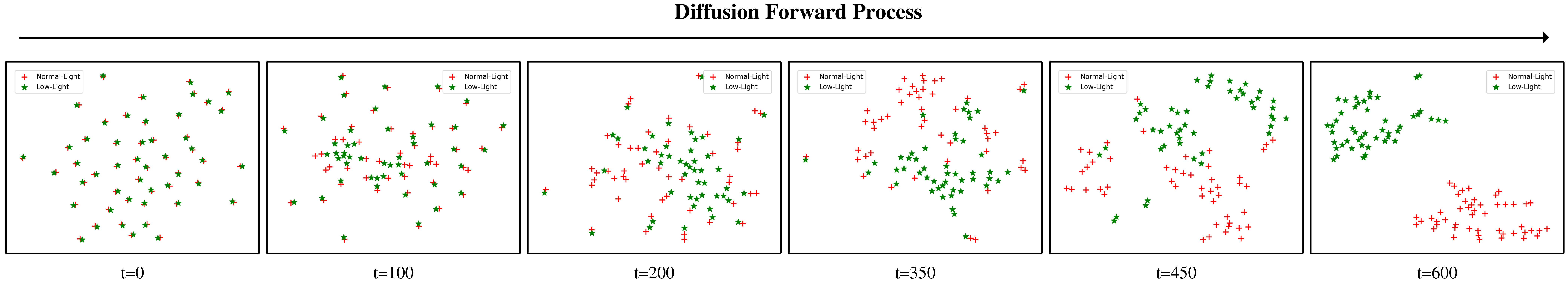}
    \caption{Distributions of the \emph{h-space} features of low-light/normal-light image pairs, as the time-step $t$ changes.}
    \label{Fig: t-SNE}
\end{figure*}

\subsection{More Discussion about the H-Space}
\label{sec: discussion about the h-space}

The diffusion models have been proven to generate images in a coarse-to-fine manner during the reverse process~\cite{choi2022perception, wang2023exploiting}, and we attribute the properties exhibited by \emph{h-space} to this.
When $t$ is small, the noise is invisibly small, and the diffusion models are focusing on processing the image content. This indicates that the \emph{h-space} features also tend to represent the background content of the image during this time. Conversely, when $t$ is large, the noise is pretty large, leading the diffusion models to concentrate on more coarse attributes. This means that the \emph{h-space} features also capture coarse features of original images, such as the foreground haze. 

This phenomenon is also prominent for other types of degradation that primarily ruins the global content of the image, such as low-light. We explore the \emph{h-space} for low-light image enhancement as the same way in Sec.4.1 of the main body. Specifically, we use the \emph{h-space} features $h^{nor}_t$ corresponding to the noisy normal-light images $y_t$ to train the decoder $D_t$. Then, we send the $h^{low}_t$ and $h^{nor}_t$ to $D_t$ to obtain the $r^{low}_t$ and $r^{nor}_t$. Note that $h^{low}_t$ denotes the \emph{h-space} feature of noisy low-light image $x_t$ at time-step $t$. We show the reconstructed $r^{low}_{t}$ and $r^{nor}_{t}$ at different time-step $t$ in Fig.~\ref{Fig: H-Space on Low-Light}. As we can see, the $r^{low}_t$ shows similar properties as $r^{haz}_t$. When $t$ is small (\ie, $t = t_1$), $r^{low}_t$ can represent the underlying content of original low-light image $x_t$. When $t$ is large (\ie, $t = t_2$), $r^{low}_t$ mainly represents the low-light characteristics of $x_t$. We also show the distribution of \emph{h-space} features using the t-SNE maps in Fig.~\ref{Fig: t-SNE}. It also shows a gradual transformation from encoding underlying content to low-light characteristics, which is similar with Fig.7 of the main body. The above observation proves that our proposed DiffLI$^2$D can be further extended to low-light image enhancement, which has been demonstrated by the superior performance in Sec.5.4 of the main body.

\subsection{More Details about Our H-Space Investigation}

\textbf{Investigation method of \emph{h-space}.}
In Sec.4.1 of the main body, we investigate the \emph{h-space} by training a decoder to map the \emph{h-space} features back to images, since the \emph{h-space} features are highly compressed and it is difficult to directly observe their properties. It seems that directly replacing the \emph{h-space} features in the denoising network $\epsilon_\theta$ is a simpler solution. For example, during the reverse sampling process of clean images, we can directly replace the \emph{h-space} features of clean images in $\epsilon_\theta$ by \emph{h-space} features of the corresponding hazy images, and then observe the sampling results to analyze the effects of \emph{h-space} features. However, we find that the direct replacement approach does not work. This is because the $\epsilon_\theta$ is usually adopted as a U-Net, and its skip-connection still injects the information of the input image into the output. Therefore, directly replacing the \emph{h-space} features will not significantly change the output of $\epsilon_\theta$, and thus the properties of \emph{h-space} cannot be effectively observed.

\begin{figure}
    \centering
    \includegraphics[width=0.70\linewidth]{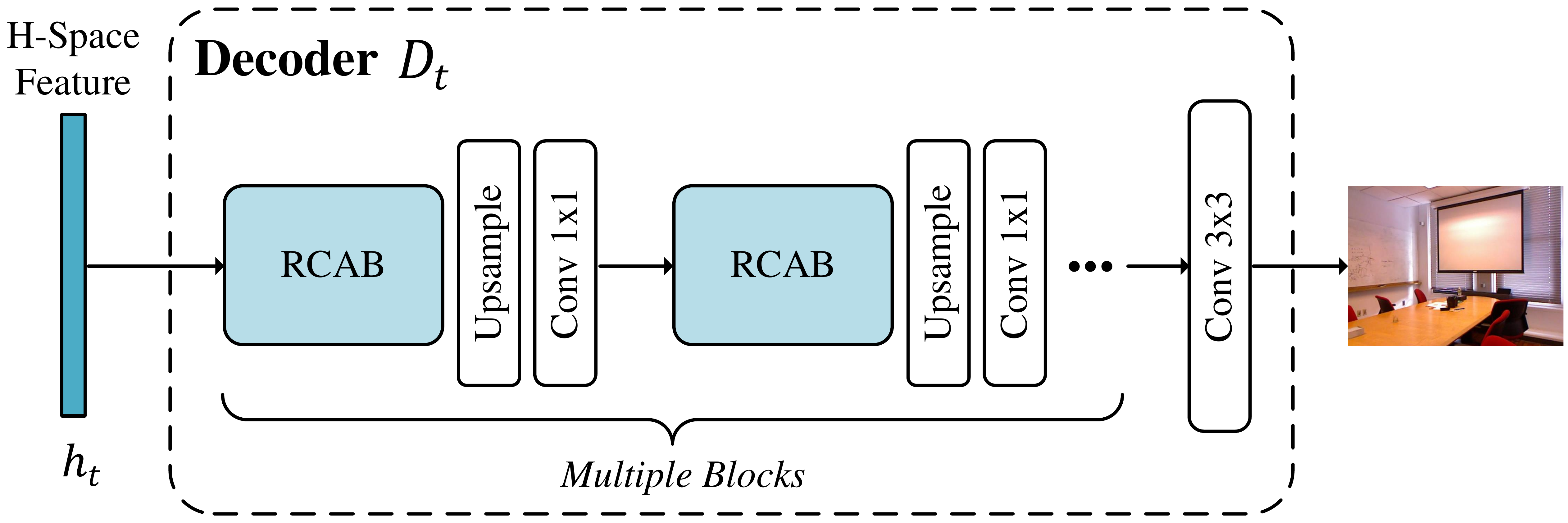}
    \caption{Architecture of the decoder $D_t$.}
    \label{Fig: Decoder}
\end{figure}

\noindent\textbf{Structure of the decoder $D_t$.}
The structure of the decoder $D_t$ is illustrated in Fig.~\ref{Fig: Decoder}. As we can see, it consists of a couple of convolution and RCAB~\cite{zhang2018image} blocks, which is lightweight and simple. Note that the decoder $D_t$ is utilized to explore the properties of \emph{h-space}, and is not involved in the training of DiffLI$^2$D.

\subsection{Comparison with Other Diffusion-Based Methods}

In this section, we compare our DiffLI$^2$D with existing diffusion model-based methods for low-level image restoration, which can be divided into three mainstream branches. The first one is to re-train a diffusion model conditioned on specific tasks from scratch~\cite{luo2023image, yi2023diff, jiang2023low, saharia2022image, wei2023raindiffusion}, and these methods tend to achieve remarkable performance. However, they typically demand massive computation resources due to the large number of parameters of the denoising network $\epsilon_\theta$, which usually has significantly more parameters than existing regression-based image restoration networks. The second approach is to directly guide pre-trained diffusion models to generate target outputs~\cite{chung2022diffusion, song2022pseudoinverse}. These methods utilize posterior sampling, and constrain the intermediate outputs during the reverse process. Compared with the first method, it avoids re-training the diffusion models. Nevertheless, they typically require time-consuming sampling to get the target image during the reverse process. On the other hand, they usually require the accurate degradation model for guidance, which cannot be obtained for image dehazing. 
The third one is to integrate diffusion models with regression-based image restoration networks~\cite{xia2023diffir, chen2023hierarchical}, which typically consists of two stages. These methods first pre-train a diffusion model to construct prior, then integrate the prior into the restoration network to train the restoration network. These methods have some similarities with ours. However, these methods require to pre-train a new diffusion model in the first stage, while the proposed DiffLI$^2$D can directly leverage the knowledge contained in the commonly-used diffusion model pre-trained on ImageNet to facilitate the image dehazing, which is more efficient.
In summary, our DiffLI$^2$D eliminates the need for re-training diffusion models, and also avoids the time-consuming reverse sampling process. Meanwhile, it effectively leverages the knowledge embedded in \emph{h-space} of the pre-trained diffusion models to facilitate the image dehazing.

\section{Additional Experiment Results}

\subsection{More Details about Datasets}

\noindent{\textbf{Low-Light Image Enhancement Datasets.}}
We evaluate our method on low-light image enhancement across LOL-v1~\cite{wei2018deep} and LOL-v2 Real~\cite{yang2021sparse} dataset. The LOL-v1 consists of 485 low-light/normal-light image pairs for training, and 15 for testing. The LOL-v2 Real contains 689 low-light/normal-light image pairs for training, and 100 for testing.

\noindent{\textbf{Deraining Dataset.}}
Rain100H~\cite{yang2017deep} is adopted as the evaluation dataset for image deraining. It consists of 1,800 rainy-clean image pairs for training, and 100 for testing.

\subsection{More Implementation Details.}
\label{sec: implementation details}
Our DiffLI$^2$D consists of two parts: the \emph{h-space} feature extraction and the image dehazing network. For \emph{h-space} features extraction, we choose the $256 \times 256$ unconditional diffusion model~\cite{dhariwal2021diffusion} as the $\epsilon_\theta$ to extract \emph{h-space} features. For the image dehazing network, we apply a 4-level encoder-decoder architecture.

\subsection{More Results about Ablation Study}

\noindent\textbf{Choice of $t$ for Other Task.} As we have demonstrated the generalization capabilities of the proposed DiffLI$^2$D on low-light image enhancement, we further provide the choice of $t$ for it. For low-light image enhancement, as discussed in Sec.~\ref{sec: discussion about the h-space}, the best choice for $t_1$ is 0. As for $t_2$, we find that the DiffLI$^2$D can achieve better performance when $t_2$ is around 600, as shown in Tab.~\ref{tab: choice of t2 for low-light image enhancement}. As we can see, the choice of $t$ for low-light image enhancement is similar with that for image dehazing. The optimal choice for $t_1$ is 0, since \emph{h-space} features can represent the underlying image content best at this time-step. And the best choice for $t_2$ is around the middle step of the diffusion process, in which the \emph{h-space} features can better represent the haze/low-light characteristics.

\begin{table}[h!]
    \centering
    \caption{Performance comparisons of different choices of $t_2$ for low-light image enhancement on LOL-v1 dataset.}
    \begin{tabular}{C{1.5cm}|C{1.0cm}C{1.0cm}C{1.0cm}C{1.0cm}}
    \shline
         $t_2$ & 200 & 400 & 600 & 800 \\
         \hline
         PSNR & 23.10 & 23.22 & \textbf{23.30} & 21.48 \\
         SSIM & 0.827 & 0.835 & \textbf{0.849} & 0.763 \\
    \shline
    \end{tabular}
    \label{tab: choice of t2 for low-light image enhancement}
\end{table}

\subsection{Evaluation on Image Deraining}
In addition to low-light image enhancement, we also test our DiffLI$^2$D on image deraining to verify the generalization capabilities. Rain100H~\cite{yang2017deep} is chosen as the evaluation dataset for training and testing. The quantitative comparison results are shown in Tab.~\ref{Tab: Image Deraining}. For comparison, we report four deraining approaches: DerainNet~\cite{fu2017clearing}, PReNet~\cite{ren2019progressive}, MPRNet~\cite{zamir2021multi}, and IR-SDE~\cite{luo2023image}. As we can see, our method achieves the best results in terms of PSNR/SSIM, and achieves comparable perceptual performance with the diffusion model-based method IR-SDE~\cite{luo2023image} in terms of LPIPS.  We also shows the qualitative results in Fig.~\ref{Fig: Deraining}.

\begin{table}[h]
    \centering
    \caption{Performance comparisons of different methods for image deraining on Rain100H~\cite{yang2017deep} dataset. The best results are marked as \textbf{bold} and the second best ones are marked by \underline{underline}.}
    \begin{tabular}{C{2.5cm}|C{1.5cm}C{1.5cm}C{1.5cm}}
    \shline
    \centering
    Method & PSNR & SSIM & LPIPS \\
    \hline
    DerainNet~\cite{fu2017clearing} & 14.85 & 0.587 & 0.573  \\ 
    PReNet~\cite{ren2019progressive} & 29.46 & 0.899 & 0.128  \\ 
    MPRNet~\cite{zamir2021multi} & 30.41 & 0.891 & 0.158  \\
    IR-SDE$^*$~\cite{luo2023image} & \underline{30.75} & \underline{0.903} & \textbf{0.048}  \\ 
    DiffLI$^2$D$^*$ (Ours) & \textbf{30.93} & \textbf{0.907} & \underline{0.053}  \\ 
    \shline
    \end{tabular}
    \label{Tab: Image Deraining}
\end{table}

\subsection{More Qualitative Results}

\begin{figure*}[h]
    \centering
    \includegraphics[width=0.98\linewidth]{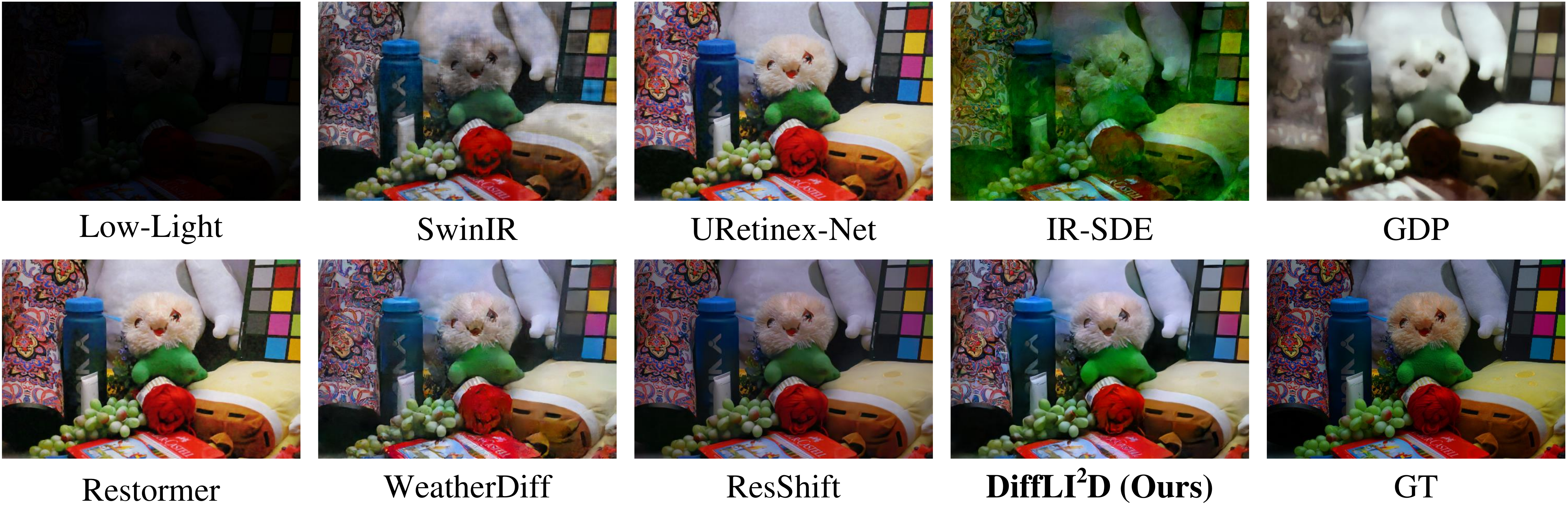}
    \caption{Qualitative results of different methods for low-light image enhancement.}
    \label{Fig: Low-Light Enhancement}
\end{figure*}

\begin{figure*}[h!]
    \centering
    \includegraphics[width=0.98\textwidth]{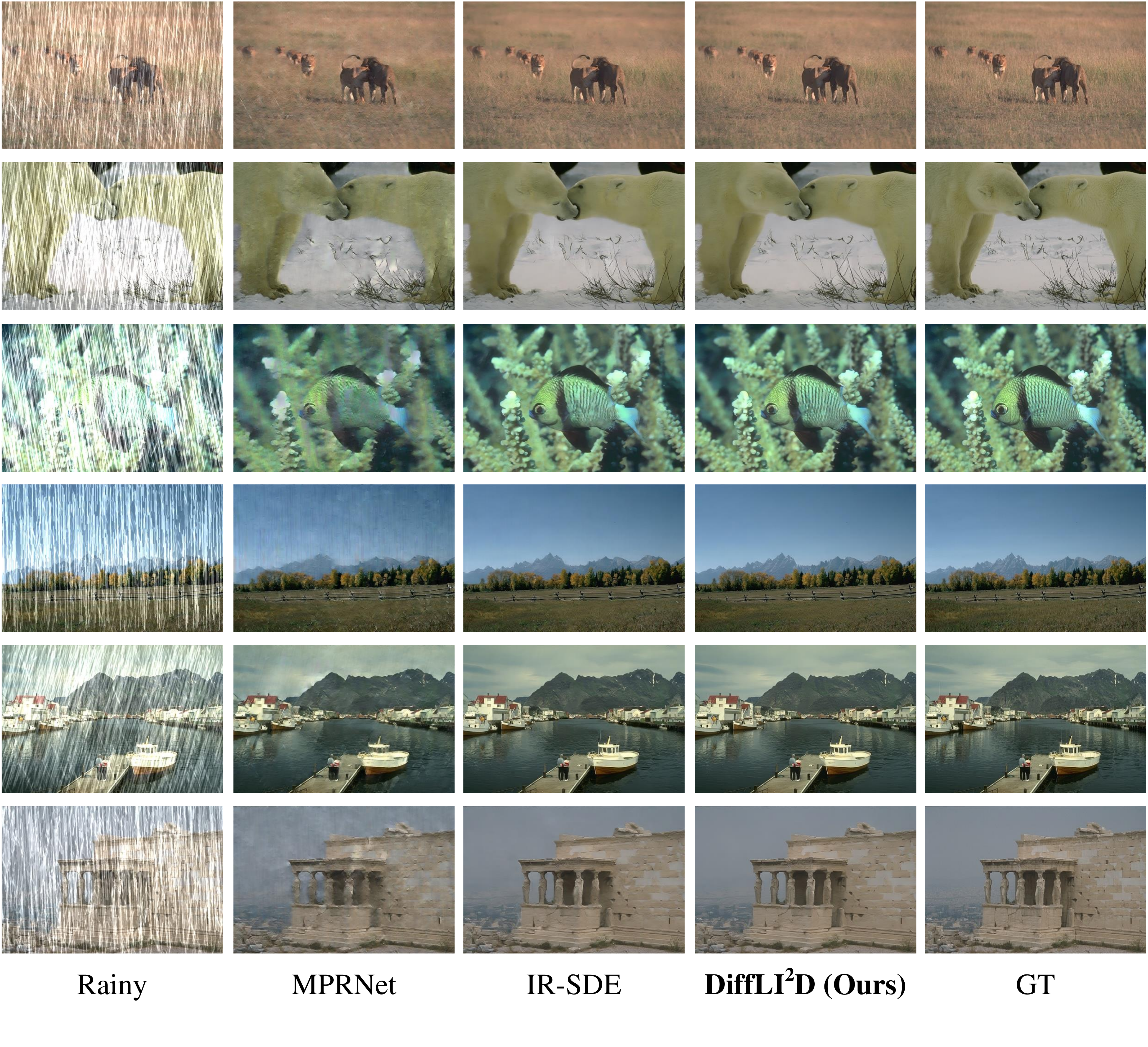}
    \caption{Qualitative results of different methods for image deraining on Rain100H dataset. Please zoom in for better view.}
    \label{Fig: Deraining}
\end{figure*}

In this section, we provide more qualitative results to demonstrate the effectiveness of the proposed DiffLI$^2$D. We further provide more qualitative results on low-light image enhancement in Fig.~\ref{Fig: Low-Light Enhancement}. And we also show the qualitative results on image deraining in Fig.~\ref{Fig: Deraining}. 

\newpage

\end{document}